\documentclass[conference]{ieeeconf}
\usepackage{graphics} 
\usepackage[pdftex]{color,graphicx}
\usepackage{algorithm}
\usepackage{algorithmic}
\usepackage{multirow}
\usepackage{epsfig} 
\usepackage{subfigure} 
\usepackage{mathptmx} 
\usepackage{balance}
\usepackage{amsmath} 
\usepackage{amssymb}  

\onecolumn
\usepackage{fancyhdr}

\usepackage{eso-pic}
\begin{document}
\AddToShipoutPictureBG*{%
  \AtPageUpperLeft{%
    \setlength\unitlength{1in}%
    \hspace*{\dimexpr0.5\paperwidth\relax}
    \makebox(0,-0.75)[c]{\Large International Journal of Robotics and Control}%
}}

\title{Robotic Supervised Autonomy: A Review}

\author{Yangming Li%
{
  \thanks{({\tt yangming.li@rit.edu})Robotic Collaboration and Autonomy Lab (RoCAL), Rochester Institute of Technology, Rochester, NY, USA 14623}%
}
}

\maketitle

\begin{abstract}
This invited paper discusses a new but important problem, supervised
autonomy, in the context of robotics. The paper defines supervised
autonomy and compares the supervised autonomy with robotic
teleoperation and robotic full autonomy. Based on the discussion, the
significance of supervised autonomy was introduced. The paper discusses the challenging and unsolved problems in
supervised autonomy, and reviews the related works in our research lab. Based on the discussions, the paper draws the
conclusion that supervised autonomy is critical for
applying robotic systems to address complicated problems in the real world.

\end{abstract}

\IEEEpeerreviewmaketitle

\pagestyle{fancy}

%

\section{Introduction}
The robotic technology meant to improve operation precision and
reliability and liberates human beings from boring, tedious, and
dangerous works. With the development of robotics, more and more
robotic systems have been applied to real-world applications and
address real-world problems. For example, industrial robots
greatly improve production precision and reliability and lower the
manufacture costs. 

One of the key factors that lead to the successful application of
robotic technology is robotic autonomy\cite{liPhd,TII18Random}. Nowadays, it generally
considers the robotic autonomy for repetitive tasks in fixed and
simple environments is a solved problem\cite{TC14Sylvester}, and robots are generally outperformed human equivalents in costs, efficiency and
reliability. However, achieving autonomy in real-world applications in
the real dynamic world remains an extremely challenging problem. This
is because, in those complicated real-world applications, robots need
to understand the environments, adapt to the environmental dynamicity,
and need to adjust the task execution automatically. Up to today, it is
still a dream to ask robots to do works for us as asking a friend.


There are many efforts we need to make, in order to push robotic
systems to the next level of applicability\cite{NN13Finite}. Among all of them, there
are some key technical challenges:
\begin{itemize}
\item sensors,

\item robotic system reliability,

\item robotic learning and intelligence,

  \item robot/human interaction.
\end{itemize}





Besides, there are also ethical/moral problems and legal concerns that
need to be addressed to introduce fully intelligent robots into our
lives. 



Some of these problems can be addressed by supervised autonomy, as it
increases the robotic system applicability by introducing human expert
knowledge and decrease the adoption barrier for robotic systems. This
paper discusses what is supervised autonomy, what are the main
differences between supervised autonomy and full autonomy and
teleoperation. It presents the key challenges and the significance of
the technology.

The paper is organized as follows: Section II defines the supervised
autonomy and compares it to full autonomy and teleoperation. Section
III introduces the significance of supervised autonomy. Section IV
discusses the key challenging problems in supervised autonomy. The
paper draws a conclusion in the last section.

\section{What is Supervised Autonomy}
\subsection{Supervised Autonomy v.s. Full Autonomy}
The final fantasy of robotics is to liberate human-being from heavy,
tedious and boring works. It is generally accepted that robotic
autonomy has different levels, which represent the increases in autonomy. There are different definitions of robotic autonomy
levels. In this paper, we focus on the relationship between autonomy and
supervised autonomy, and pick the autonomy definition for autonomous
cars, as a reference, to facilitate the discussion.

For autonomous cars, there are 5 levels of autonomy, and level 5,
is considered full autonomy. On the bottom of the 5 levels, it is
no autonomy at all, we consider it as level 0.

\subsubsection{Level 0: No Autonomation}
Electrical/Mechanical systems extend human capabilities. These systems
are fully controlled by human-being and do not assist human with the
operations. The traditional cars are such examples. These systems, no
matter how complex they are, they are intrinsically tools, and fully
operated by human-being.

\subsubsection{Level 1: Robotic Assistance Without Environmental Perception}
This level of autonomy is primitive. The systems with such autonomy
can assist human with operations, often based on human's setup and
simple rules. For example, traditional cruise control. When
drivers set up the cruise speed, a car can use its speed sensor to
close the control loop and ease human drivers from the tedious gas
pedal control.
 
\subsubsection{Level 2: Robotic Assistance With Environmental Perception}
Level 2 autonomy still depends on simple rules, however, such robots
can actively perceive environments and adapt to environmental changes,
to a certain degree. For example, level 2 autonomous cars can react
to traffic lights. 
\subsubsection{Level 3: Robotic Operation Under Human Monitoring}
When robots reach level 3, they can continuously monitor environments,
and autonomously adjust operations based on environmental
changes. However, robots are still not capable of adapt to all
conditions and will require human's monitoring all the time. 

\subsubsection{Level 4: Robotic Operation With Human Supervision}
The level 4 autonomy, for autonomous cars, is fully autonomous, but
need supervision. Level 4 autonomous cars can adapt to various
road conditions and environments, and only need human decisions for
rare cases. 

\subsubsection{Level 5: Full Autonomation}
The level 5 autonomy, for autonomous cars, is the equivalent of human
drivers. These robots behave as taxi drivers and do not need human
intervention.


From the upper discussion, it is clear that robots with full
autonomy are the equivalent of human experts. It worth to point out
that, compared to driving, most of the real world tasks are more
complicated. Human often needs days to learn to drive, but a lot of
professional services, such as surgery, takes years of training and
practice the rules are not straightforward. It will be much harder
to realize full autonomy in many fields.


\subsection{Supervised Autonomy v.s. Teleoperation}
Teleoperation\cite{teleoperation} refers to the remote control of a
robot from a console. In a teleoperated robotic system, a human
operator controls the movements of the slave side of a robot. It is
clear that teleoperation is opposite to full autonomy as human
operators fully control the robot. 

\subsection{Levels of Supervised Autonomy}
From the upper discussion, we can see that teleoperation solely depend
on human micro-level control for performing tasks, and controlling
robot behavior. Autonomy, as a comparison, purely rely on robots on
task performing. However, in real-world applications, operations and
behavior are under supervision. The more dangerous the operations are,
the more intense and more complicated supervision will be
conducted. Despite the reasons from the legal perspective, there are
real technical motivations. Expert knowledge is needed to ensure the
successful execution, reliability, and the results meet
expectations. From this perspective, autonomy is always supervised. 
 
\begin{figure}
  \centering
      \includegraphics[height=0.3\textwidth]{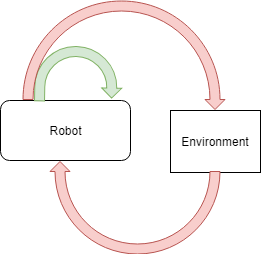}
      \caption{Perception in Supervised Autonomy. In low-level supervised autonomy, robots close the loop of control by monitoring self-status. In high-level supervised autonomy, robots have environmental perception and can perform complex tasks. }
    \label{fig:supervisedAutonomy}
  \end{figure}

Being similar to full autonomy, supervised autonomy has different
levels. These levels reflect the increase of the capability of
environmental perception, and the capability to solve tasks with
increasing complexities\cite{campbell1988task} Fig. \ref{fig:supervisedAutonomy}.

\subsubsection{Level 1: Robotic Assistance}
This is the lowest level of supervised autonomy. In this level of
supervised autonomy, human setup the goal status for robots, and
robots adjust self status.

This level of supervised autonomy liberate human from simple repetitive
micro controls. But the robots can neither adapt to environmental
changes nor perform complex tasks. 
 

\subsubsection{Level 2: Entry-Level Task Autonomy Under Supervision}
Robots with level 2 supervised autonomy can perform entry-level tasks,
under continuous and intensive supervision from human-being. These robots
have the superior capability on environmental perception and task
execution, however, lack the capability of complex task planning and
decision making. 

Task complexity is a well-studied problem and there are many
definitions\cite{campbell1988task}. In the context of supervised
autonomy study, we categorize tasks into two categories, the entry-level tasks, and the specialist-level tasks. The entry-level tasks are
based on simple rules, and the results can be objectively measured by
robots. For example, when a robot grasps an object, the rules for the
task is explicit and simple, and the results can be objectively and
simply measured by the robot.


\subsubsection{Level 3: Specialist-Level Task Autonomy Under
  Supervision}
Level 3 supervised autonomy still requires continuous supervision from
human-being. Being different with robots at level 2, robots at level 3
reach human level capability of task execution, and can perform
complex tasks autonomously under human supervision. 

Note that raising from level 2 to level 3, robots
need not only improved intelligence but also improve hardware and
system. This is because the execution of more complex tasks requires
robots having improved capability, performing collaborations, and
having superior intelligence. 

\subsubsection{Level 4: Autonomy With Human Supervision}
Robotic systems with level 4 supervised autonomy have
human-expert-level reliability. These robotic systems no longer need
intensive supervision from human experts. Instead, they behave as human experts
and actively consult human experts if the systems think it is needed.

\section{Why Supervised Autonomy?}
Supervised autonomy is needed because human experts' opinions are
the gold standard. Even through supervised autonomy is closely related to
robotic teleoperation and full autonomy, it has clear
differences. When robotic systems are teleoperated, the systems
require no autonomy, the control is fully on human-being and
supervised autonomy is not needed. 
When robotic systems are fully autonomous, the control is fully
independent of human control, and supervised autonomy is not needed.



\subsection{Robotic System Reliability}
Robotics systems have many components, and a single problem in any of
these components can jeopardize the performance and the stability of
robots\cite{JU13KMP,ICRA16Hysteresis,li2008particle2,li2007quadtree}. Improving the stability and the performance of robotic
systems requires the development of robotic technology and the
accumulation of experiences\cite{haghighipanah2016unscented,guo2008}. Moreover, improving system reliability
often relies on extra hardware\cite{gateway,arm,li2005} and software\cite{chen2007auto,li2007sp,chen2007energy}, which further improve the
complexity of robotic systems\cite{li2012distributed,chen2008error,ICRA17Endoscope,chen2008localization,yang2008wsn,li2008particle,meng2007localization}.

Robotic systems continuously work in dynamic real environments\cite{NCAA18STMVO,aghdasi2015atlas,Neurocomputing13SLAMIDE}, which
has various adverse factors that cause a robotic system failure\cite{haghighipanah2015improving}. Because
of the system complexity, the environmental complexity, and the task
complexity, it is extremely challenging to maintain robots'
performance and stability in real-world applications, under today's
technology, regardless of the improvement of robotic technology and
researchers' efforts. Experts' supervision on robotic systems can
serve as guardians to robotic systems and allows introducing robotic
systems into real-world applications. 



\subsection{Robotic Intelligence}
Most of the existing robotic research focuses on robotic
technology\cite{TNNLS13WTA,li2012using}. However, domain knowledge is needed in many real-world
applications\cite{JNSB17Region,JAMA19Nasal,JNSB17AnatomicalRegion}. Expert level knowledge is essential to successful
robotic applications but is difficult to achieve\cite{JNSB17Atlas,JNSB18RelativeMotion,JNSB18Completeness}. Classical expert
systems utilize rule-based intelligent systems and facing the
exponential complexity increases\cite{SI17QuantitativeMotion,JNSB17AutomatedAssess}, thus are not easy to develop and
maintain for complex applications\cite{hannaford2016simulation,harbison2016objective}. Recently, deep learning based
methods achieved impressive progress and outperform human performance
in many applications\cite{PLoS19,Frontiers19Protein}, such as natural language processing. However,
deep learning methods often require a large amount of training data\cite{ICRA19Segmentation},
which is often not available for robotic applications. Moreover, deep
learning methods are often applied to address single-task problems and
are sensitive to the change of data distributions. As a result,
equipping robots with expert level skills is still a challenging and
unsolved problem. Because of these limitations, supervised autonomy
plays a significant role in accelerating introducing robots into
real-world applications.


\subsection{Robotic Collaboration}
Collaboration is essential for complex tasks, even for a human
being. Although there is a large amount of existing research and effort
towards robot/human collaboration\cite{TNNLS17Cooperative}, existing results often aim to
address single-task applications\cite{TII18Cooperative}. Therefore, utilizing human expert
knowledge to decompose and simplify tasks into simpler tasks, which
can be handled by robots, is important to extend robotic applications.


\subsection{Ethical and Legal Vacancy}
Although the study in robotics has made impressive progress in the
past to centuries, and robots already started to address real-world
problems, it is still blank in ethics and law for robots directly
interacting with a human. For simple operations, such as driving, human
beings can often reach a common consensus on the evaluation of the
operations. For example, an operation causes a traffic accident and
damage to human or properties are definitely failed. For complex
operations, such as surgeries, even human experts can have conflict
opinions toward some operations. This is often caused by the fact that
the evaluation of the operation is complicated. as the results, the
evaluations of hypothesizing operations are more complicated and often
controversial. When the huge loss of values is associated with
operations, human appeal lawsuit to seek for solutions. When such
situation raised, a committee formed by human experts will be the
reference for the court. For robots, when concerns are raised
regarding the operations, human experts will evaluate the
results. Clearly, supervised autonomy allows human experts to make
critical decisions to ensure the robotic systems are safe and
effective.  



\section{Challenges and Opportunities in Supervised Autonomy}

\subsection{Sensor Information}
Sensors are fundamental and challenging components in robotic
systems\cite{HU13LIDAR,Sensors13LIDAR,ICRA11StructureTensor,ICRA10LIDAR}. Even after decades of research, it is still difficult to
increase robot perception to human level\cite{li2008visual}. For example, tactile sensing
is essential and fundamental to a human being but despite the
impressive progress of force sensor matrix research, robot haptic
sensing is far away from human performance\cite{haptic2017}, in both sensing precision
and resolution\cite{RAL17GPR,ICRA16DynamicModel}.

Another important problem is sensor fusion\cite{li2013,MP17Segmentation,JMI17Segmentation}. Human naturally uses all available information\cite{li2013perception,li11}, such as vision, hearing, and tactile, for performing tasks, but for robots, sensor fusion is still a challenging problem, especially with comparatively poor sensor information quality\cite{TII14IPJC,IROS12IPJC,li2007data}.


\subsection{Robot Control}
Robot control is a historical problem and remains challenging and attractive. There is a huge research community, which focuses on improving control efficiency\cite{li2012model} and system robustness\cite{IROS18RNNSoft,IROS17RNNAdaptiveness}. However, while modern robots often have redundancy for improved system reliability, it makes the control problem harder\cite{ICRA18RNNPlan}. While multiple redundant robots work collaboratively, the control problem reaches a new level of complexity\cite{li2012decentralized2,li2012decentralized}. 

\subsection{Robotic Intellgence}
Human intelligence keeps increasing from new experiences and is superior in heuristic learning and reasoning\cite{JOHNS17FreeFlap}. Robots need a similar learning capability to keep increasing performance. Robots also need to increase the reasoning capability to transform knowledge from domains to domains, which is a problem need to be addressed as soon as possible. 


\subsection{Robot/Human Interaction and Interface}
Supervised autonomy needs robot-human interaction. The classical robot/human interaction has insufficient efficiency for supervised autonomy\cite{ICSM19Collaboration}. We need more than emergency stops to guide robots for improving robotic system performance\cite{THC15EEG}. 

\section{Conclusion}
This work introduces supervised autonomy, a new and important topic in the context of
robotics. The paper defines supervised autonomy, and through comparing with
teleoperation and full autonomy, explains the significance and the
importance of this topic. The paper discusses challenges and
opportunities in Supervised Autonomy, and hope to help other
researchers to quickly push forward the development of supervised autonomy.

\section*{Acknowledgment}

\balance
 \newcommand{\noop}[1]{}

\end{document}